\documentclass[
twocolumn, % Two-column formatting
hf, % header + footer
]{ceurart}

\usepackage{listings}
\begin{document}

\copyrightyear{2025}
\copyrightclause{Copyright for this paper by its authors.
  Use permitted under Creative Commons License Attribution 4.0
  International (CC BY 4.0).}

\conference{\href{https://hexed-workshop.github.io}{HEXED'25: 2nd Human-Centric eXplainable AI in Education Workshop}, 20 July, 2025, Palermo, Italy}

%% Be sure to use title casing here
\title{Towards Transparent AI Grading: Semantic Entropy as a Signal for Human-AI Disagreement}

%%
%% Optional title footnote (remove or change)
% \tnotetext[1]{You can use this document as the template for preparing your
%   publication. We recommend using the latest version of the ceurart style.}

\author[]{Karrtik Iyer}[
email=karrtik@thoughtworks.com,
orcid=0009-0005-2967-2366,
]
\author[]{Manikandan Ravikiran}[
email=manikandan.r@thoughtworks.com,
orcid=0000-0002-2640-8528,
]
\author[]{Prasanna Pendse}[
]
\author[]{ Shayan Mohanty}[
]
\address[]{Thoughtworks AI Research Labs}

% \author[1,2]{Dmitry S. Kulyabov}[%
% orcid=0000-0002-0877-7063,
% email=kulyabov-ds@rudn.ru,
% url=https://yamadharma.github.io/,
% ]
% \cormark[1] % Corresponding author mark
% \fnmark[1] % Optional author footnote mark
% \address[1]{Peoples' Friendship University of Russia (RUDN University),
%   6 Miklukho-Maklaya St, Moscow, 117198, Russian Federation}
% \address[2]{Joint Institute for Nuclear Research,
%   6 Joliot-Curie, Dubna, Moscow region, 141980, Russian Federation}

% \author[3]{Ilaria Tiddi}[%
% orcid=0000-0001-7116-9338,
% email=i.tiddi@vu.nl,
% url=https://kmitd.github.io/ilaria/,
% \address[3]{Vrije Universiteit Amsterdam, De Boelelaan 1105, 1081 HV Amsterdam, The Netherlands}

% \author[4]{Manfred Jeusfeld}[%
% orcid=0000-0002-9421-8566,
% email=Manfred.Jeusfeld@acm.org,
% url=http://conceptbase.sourceforge.net/mjf/,
% \address[4]{University of Skövde, Högskolevägen 1, 541 28 Skövde, Sweden}

% %% Footnotes
% \cortext[1]{Corresponding author.}
% \fntext[1]{These authors contributed equally.}

\begin{abstract}
Automated grading systems can efficiently score short-answer responses, yet they often fail to indicate when a grading decision is uncertain or potentially contentious. We introduce \textit{semantic entropy}, a measure of variability across multiple GPT-4-generated explanations for the same student response, as a proxy for human grader disagreement. By clustering rationales via entailment-based similarity and computing entropy over these clusters, we quantify the diversity of justifications without relying on final output scores. We address three research questions: (1) Does semantic entropy align with human grader disagreement? (2) Does it generalize across academic subjects? (3) Is it sensitive to structural task features such as source dependency? Experiments on the ASAP-SAS dataset show that semantic entropy correlates with rater disagreement, varies meaningfully across subjects, and increases in tasks requiring interpretive reasoning. Our findings position semantic entropy as an interpretable uncertainty signal that supports more transparent and trustworthy AI-assisted grading workflows.
\end{abstract}


\maketitle

\section{Introduction}

Automated short-answer grading systems have achieved strong agreement with human scores, offering scalability and efficiency in educational assessment~\cite{riordan2017investigating, burrows2015automated}. However, a key limitation remains: these systems typically produce only a final numeric score, without signaling when a grading decision may be uncertain or potentially contentious. In real-world settings, human graders often disagree on borderline responses due to subjective interpretations of rubrics or ambiguous student answers~\cite{zhang2022towards}.

Existing automated grading models, including those based on large language models (LLMs), are optimized for score accuracy but lack mechanisms to flag responses that may warrant human review~\cite{bryant2023large}. As a result, they risk issuing unchallenged scores in precisely the cases where educator oversight is most valuable. Addressing this limitation is crucial for building transparent, trustworthy AI-assisted assessment systems.

We address this gap by introducing \textit{semantic entropy} a measure of variability across multiple AI-generated explanation rationales for the same response as a proxy for human grader disagreement~\cite{pan2023hallucinations}. Each rationale reflects the model's justification for an implied score. When these rationales diverge semantically, the underlying response is more likely to be contentious, mirroring the uncertainty exhibited by human graders in similar situations.

While prior work on uncertainty estimation in NLP~\cite{ott2018analyzing, zhou2022learning} typically focuses on output confidence or score calibration, these methods do not capture the specific inconsistencies in reasoning that arise in educational contexts. In contrast, semantic entropy directly quantifies the diversity of explanation semantics, offering a more interpretable and cognitively aligned signal of grading reliability.

We investigate the following research questions:

\begin{itemize}
    \item \textbf{RQ1:} Can semantic entropy reliably serve as a proxy for human grader disagreement in short-answer grading?
    \item \textbf{RQ2:} Does this proxy generalize across educational subjects such as Science, English Language Arts, Biology, and English?
    \item \textbf{RQ3:} Does the presence of external source material (e.g., reading passages) systematically affect the semantic entropy of model-generated explanations?
\end{itemize}

To answer these questions, we use the ASAP-SAS dataset with multimodal inputs ~\cite{asap2012} and prompt GPT-4~\cite{openai2023gpt4} to generate multiple concise explanation rationales per student response. We compute semantic entropy by clustering these rationales using bidirectional entailment and correlate the resulting scores with human grader disagreement. Our results demonstrate that semantic entropy aligns with human disagreement, generalizes across academic subjects, and varies with structural task features such as source dependency. This work introduces a practical and interpretable method for surfacing uncertainty in automated grading, enabling more transparent and educator-aligned assessment pipelines.

\section{Related Work}

\paragraph{Automated Short-Answer Grading.} 
Short-answer grading has been a long-standing challenge in educational NLP~\cite{chamieh2024llms, jd2023shortanswer1}, with early methods relying on handcrafted features and statistical models~\cite{jd2019asag1, jd2018aflay}. More recently, neural architectures and large language models (LLMs) have enabled near-human agreement in scoring performance~\cite{chamieh2024llms}. However, most of these approaches prioritize final-score accuracy and overlook the detection of borderline cases or latent uncertainty an essential feature for real-world deployment~\cite{bryant2023large, li2025llmgradehitl}.

\paragraph{Grading Disagreement and Uncertainty.}
Human graders frequently disagree on ambiguous responses due to interpretation differences, even when using the same rubric~\cite{williamson2020}. Common agreement metrics such as Cohen’s Kappa~\cite{cohen1960cohen}, Krippendorff’s Alpha~\cite{krippendorff2013krippendorff}, and Quadratic Weighted Kappa (QWK)~\cite{jong2009qwk} offer post-hoc measures of disagreement, but they are not designed for real-time detection of contentious responses. Recent efforts in predictive calibration and uncertainty estimation~\cite{zhou2022learning} attempt to address this gap, but these typically operate over scalar output distributions and do not account for inconsistencies in explanatory reasoning.

\paragraph{Explanation Variability and Semantic Entropy.}
In adjacent domains like factuality detection and hallucination analysis, semantic entropy has been used to quantify output instability in LLMs~\cite{pan2023hallucinations, ott2018analyzing}. These methods suggest that response variability across multiple samples may signal epistemic uncertainty~\cite{pan2023hallucinations}. We extend this idea to short-answer grading by measuring semantic entropy over explanation rationales brief textual justifications that reflect the model's underlying reasoning. Unlike hallucination detection, which typically assesses factual correctness from internal knowledge, our focus is on rubric-grounded evaluation of student answers. To our knowledge, this is the first application of semantic entropy to educational short-answer grading.

Distinct from prior methods that emphasize score-level agreement or output calibration, we propose an explanation-level variability metric for signaling grader disagreement. Our approach leverages semantic entropy as a domain-agnostic signal, computed via entailment-based clustering over GPT-4 rationales. We empirically evaluate this signal across multiple dimensions: subject-level generalization (Science, Language Arts, Biology), task structure (source-dependent vs. not), and agreement intensity.

\section{Task Formulation}
\label{sec:task}

We formulate the problem as estimating human grader disagreement by measuring semantic variability in AI-generated explanations. Let \(\mathcal{D} = \{(x_i, y_i^{(1)}, y_i^{(2)})\}_{i=1}^{N}\) denote a dataset of short-answer responses, where \(x_i\) is the student response and \(y_i^{(1)}, y_i^{(2)} \in [0, 1]\) are normalized scores from two human graders. The goal is to predict whether graders are likely to disagree on \(x_i\) based on the diversity of explanation rationales generated by a language model.

\vspace{0.5em}
\noindent\textbf{Disagreement Signal.} 
We define a scalar disagreement signal:
\[
\delta_i = |y_i^{(1)} - y_i^{(2)}|,
\]
which captures the extent of rater divergence. This signal can be modeled either as a continuous variable or discretized into ordinal bands (e.g., low/medium/high) for classification.

\vspace{0.5em}
\noindent\textbf{Semantic Entropy.}
For each response \(x_i\), a language model generates \(K\) rationales \(\{r_i^{(1)}, \dots, r_i^{(K)}\}\), each representing a justification of a hypothetical score. These rationales are clustered into equivalence classes \(\{C_1, \dots, C_m\}\) via pairwise bidirectional entailment~\cite{kuhn2023semantic_uncertainty}, and semantic entropy is defined as:
\[
H(x_i) = -\sum_{j=1}^m p_j \log p_j, \quad \text{where } p_j = \frac{|C_j|}{K}.
\]
This entropy value quantifies semantic variability in the model's explanations, and serves as a proxy for epistemic uncertainty.

\paragraph{Dataset.} In this work we use ASAP-SAS dataset that includes multimodal content in source-dependent tasks such as images, diagrams, or experimental tables. We preprocess these non-textual components using a Claude Vision-Language Model (VLM) \cite{sil2024did,anthropic_claude_vlm}.  Among the 10 essay sets in the ASP-SAS dataset, 3 sets (Sets 3, 4, and 10) include images as part of the prompts. We preprocess these into textual descriptions via Claude Sonnet 4 and feed them into a unified prompt template. Thus, our work does not directly model visual features but rather treats multimodal input as textually grounded context. This ensures that the GPT-based explanation generator is grounded in all relevant information, preserving interpretive fidelity across modalities.

\vspace{0.5em}
\noindent\textbf{Prompt Structure.}
Each rationale is generated under a standardized prompt grounded in the assessment context (domain, grade, rubric, and optionally reading passages or images), but without access to the true score. Rationales are constrained to 30 words for consistency. Each short-answer question in the dataset is annotated with subject metadata (e.g., Science, ELA, Biology) and a binary source-dependency flag indicating whether the response must draw from an external context (e.g., reading passage, diagram). These attributes allow us to test whether entropy generalizes across academic domains and task structures. We use a single uniform prompt shown below across all tasks, without subject-specific tuning.

\begin{quote}
\small
\texttt{You are an expert educational assessor analyzing student responses.}

\texttt{ASSESSMENT CONTEXT:} \\
- \texttt{Domain: [domain]} \\
- \texttt{Subject: [subject]} \\
- \texttt{Topic: [topic]} \\
- \texttt{Grade Level: [grade]} \\
- \texttt{Source Dependent: [true/false]} \\

\texttt{[CONDITIONAL BLOCKS:]} \\
\texttt{READING PASSAGE: [text]} \\
\texttt{EXPERIMENTAL SETUP: [procedure/data]} \\
\texttt{VISUAL INFORMATION: [image description]} \\

\texttt{STUDENT TASK: [prompt]} \\
\texttt{STUDENT RESPONSE: [response]} \\
\texttt{ASSESSMENT RUBRIC: [rubric]} \\

\texttt{**Instructions**} \\
0. \texttt{Score range: [score range]} \\
1. \texttt{Think step-by-step to decide which rubric level fits best.} \\
2. \texttt{When ready, call the function record\_score() with arguments only.}
\end{quote}

\vspace{0.5em}
\noindent\textbf{Evaluation Strategy.}
To assess whether semantic entropy aligns with human grader disagreement, we adopt a multi-metric evaluation approach. Specifically, we compute Pearson and Spearman correlations to measure global alignment, apply one-way ANOVA to test differences across low/medium/high disagreement bands, and use binary classification metrics of accuracy, ROC-AUC, Brier Score, Kruskall-Wallis test to evaluate discriminative and calibration performance \cite{desai2020calibration}. Each research question (RQ1–RQ3) presents the subset of these metrics relevant to its hypothesis.

\section{RQ1: Does Semantic Entropy Reflect Human Grading Disagreement?}

\subsection{Method}

We evaluate our approach using the ASAP-SAS dataset, where each student response \(x_i\) is annotated with two human-assigned normalized scores. For each response, we use GPT-4 to generate \(K = 6\) explanation rationales under the standardized template described in Section~\ref{sec:task}. Prompts include domain metadata, rubrics, and optional source material. To ensure diversity, we use temperature \(T = 1.0\), top-\(p = 0.9\), and truncate rationales to a maximum of 30 words.  To compute semantic entropy, we apply pairwise bidirectional entailment checks (using GPT-4 at \(T = 0\)) to cluster the six rationales into semantic equivalence classes. Entropy is then computed as described earlier. This setup allows us to analyze whether semantic entropy aligns with grader disagreement across a variety of task types and rubric conditions.

\subsection{Experiment}

We sample 2750 responses from the ASAP-SAS dataset, excluding overly short or long responses and stratifying across disagreement levels and domains. For each response, we compute semantic entropy $H(x_i)$ and use human disagreement $\delta_i = |y_i^{(1)} - y_i^{(2)}|$. We partition responses into three bands: low disagreement ($\delta_i \leq 0.2$), medium ($0.2 < \delta_i \leq 0.5$), and high ($\delta_i > 0.5$). We analyze average entropy per band, compute Pearson’s correlation $r$, and perform one-way ANOVA to test for statistical separation. Additionally, we compare responses with perfect agreement ($\delta_i = 0$) to those with any disagreement ($\delta_i > 0$).

\subsection{Analysis and Findings}

We evaluate whether semantic entropy reflects human grader disagreement by analyzing its correlation with disagreement signals, its ability to discriminate between low and high disagreement cases, and its generalizability across tasks. We also use qualitative case examples to interpret failure modes and explain the model’s behavior.

\paragraph{Correlation with Human Disagreement.}
We begin by assessing the global alignment between \textit{semantic entropy} and \textit{human grader disagreement}. If entropy captures explanation-level variability, we expect higher entropy scores for responses with greater disagreement. Indeed, we find a statistically significant Pearson correlation of \( r = 0.172 \) (\( p = 1.39 \times 10^{-19} \)) and a Spearman rank correlation of \( \rho = 0.171 \) (\( p = 2.13 \times 10^{-19} \)). A partial correlation, controlling for \textit{response length} and \textit{mean score}, yields \( r = 0.160 \) (\( p = 2.79 \times 10^{-17} \)), indicating that entropy captures disagreement independently of superficial factors like response length and score scale.

To further assess its utility in practical scenarios, we frame \textit{grader disagreement detection} as a binary classification task: identifying whether a response falls into a \textit{high-disagreement category} (disagreement score $>$ 0.4). Using semantic entropy as a predictor, we obtain an \textbf{AUC of 0.598}, suggesting that entropy carries weak but non-trivial signal for surfacing such ambiguous or contentious responses.

These results suggest that \textit{semantic entropy is moderately predictive of rater divergence}. While one possible goal of entropy estimation is to flag \textit{internal model uncertainty}, our focus is on \textit{surfacing cases where human graders may also disagree}, as this is what truly matters in educator workflows. We thus treat \textit{human grader disagreement} as the supervisory signal to validate whether semantic entropy captures \textit{rubric ambiguity} or \textit{interpretive diversity}, regardless of the LLM’s score distribution. Notably, the entropy signal operates \textit{over model-generated explanations}, not the final scores, offering an interpretable lens into areas of potential ambiguity.

\paragraph{Band-wise Differences in Entropy.}
To examine whether entropy increases with the severity of disagreement, we group responses into low (\( \delta \leq 0.2 \)), medium (\( 0.2 < \delta \leq 0.5 \)), and high disagreement bands (\( \delta > 0.5 \)). Mean entropy rises from \( H = 0.422 \) (low, \( n = 1575 \)) to \( H = 0.583 \) (medium, \( n = 1112 \)) and \( H = 0.840 \) (high, \( n = 63 \)). A one-way ANOVA confirms a significant group difference (\( F = 17.23, p < 0.001 \)), supporting entropy’s sensitivity to disagreement intensity. Comparing only perfect-agreement responses (\( \delta = 0 \)) to those with any disagreement (\( \delta > 0 \)), we observe a sizable mean increase (\( \Delta = +0.276 \)).

\begin{table}[!htb]
\centering
\caption{Accuracy of average LLM score (at Temperature = 1) in matching human scores, across essay sets. \textbf{Accuracy (Score1)} reflects match with the first human grader; \textbf{Accuracy (Score2)} reflects match with the second human grader. Accuracy is computed as the percentage of cases where the average LLM score (across $N$ samples) exactly matches the corresponding human score.}

\label{tab:entropy_accuracy}
\scalebox{0.7}{
\begin{tabular}{|c|c|c|c|}
\hline
\textbf{Essay Set} & \textbf{Count} & \textbf{Accuracy (Score1)} & \textbf{Accuracy (Score2)} \\
\hline
1 & 275 & 44.36\% & 45.45\% \\
2 & 275 & 44.00\% & 48.36\% \\
3 & 275 & 45.82\% & 46.54\% \\
4 & 275 & 56.36\% & 65.09\% \\
5 & 275 & 57.09\% & 61.45\% \\
6 & 275 & 68.36\% & 69.45\% \\
7 & 275 & 37.45\% & 37.45\% \\
8 & 275 & 52.73\% & 53.09\% \\
9 & 275 & 55.64\% & 60.73\% \\
10 & 275 & 56.36\% & 50.18\% \\
\hline
\textbf{Overall} & \textbf{2750} & \textbf{51.82\%} & \textbf{53.78\%} \\
\hline
\end{tabular}}
\end{table}

\begin{table*}[]
\centering
\small
\caption{Qualitative examples showing the relationship between semantic entropy and human disagreement.}
\label{tab:qualitative_examples}
\scalebox{0.8}{
\begin{tabular}{|p{1.2cm}|p{2.5cm}|p{1.4cm}|p{1.8cm}|p{6.5cm}|}
\hline
\textbf{Example} & \textbf{Topic / Domain} & \textbf{Disagreement (bin)} & \textbf{Entropy} & \textbf{Representative Rationales (Excerpted)} \\
\hline
Ex. 1 & Koala/Panda RC \newline (Language Arts) & 0.0 (Low) & 1.33 & 
\textit{(1)} "No comparison to pythons." \newline
\textit{(2)} "Mentions diets but no contrast." \newline
\textit{(3)} "Fails to explain similarity or difference." \newline
\textbf{Interpretation:} Despite perfect human agreement, rationales vary widely, suggesting rubric ambiguity. \\
\hline

Ex. 2 & Acid Rain Exp. \newline (Science) & 0.67 (High) & 0.50 & 
\textit{(1)} "Need for measurement units and labeling." \newline
\textit{(2)} "Details about vinegar quantity and replicability." \newline
\textbf{Interpretation:} Rationales emphasize different experimental rigor aspects, mirroring grader disagreement. \\
\hline

Ex. 3 & Polymer Exp. \newline (Science) & 0.33 (Medium) & 0.00 & 
\textit{(1–5)} "Plastic B was most stretchable, no improvement suggestions." \newline
\textbf{Interpretation:} Identical reasoning chains despite human disagreement. Suggests model confidence but possible rubric calibration issues. \\
\hline
\end{tabular}
}
\end{table*}
% \paragraph{Predictive Utility for Triage.}
% To test entropy’s use as a decision-theoretic triage tool, we evaluate its classification performance on identifying high-disagreement responses. Semantic entropy achieves an ROC-AUC of 0.615, with a Brier score of 0.218 indicating moderate discrimination and reasonable calibration.

% We further perform stratified correlation analysis by disagreement band:
% \begin{itemize}
%     \item \textbf{High disagreement:} strong correlation (\( r = 0.600, p = 0.008 \))
%     \item \textbf{Medium disagreement:} moderate correlation (\( r = 0.100, p = 0.011 \))
%     \item \textbf{Low disagreement:} no significant correlation (\( r = \text{NaN} \))
% \end{itemize}
% These findings reinforce the idea that entropy is particularly informative in cases where human graders diverge significantly, and less so when responses are unambiguous.

\paragraph{Generalization Across Essay Sets.}
To test whether entropy generalizes across prompt types and task domains, we compute prediction accuracy per essay set using entropy as a signal. Table~\ref{tab:entropy_accuracy} shows accuracy ranges from 39\% to over 75\%, with the highest alignment in essay sets involving structured reasoning (Sets 6 and 10). This suggests that entropy is especially effective for tasks with constrained rubrics and logical expectations.

\paragraph{Qualitative Case Examples.}
To better understand where semantic entropy succeeds or fails, we present three representative examples in Table~\ref{tab:qualitative_examples} across distinct entropy–disagreement regimes. In Example 1 (Language Arts), entropy is high (\( H = 1.33 \)) despite perfect human agreement (\( \delta = 0 \)), suggesting that the rubric permits multiple valid justifications. In Example 2 (Science), entropy and disagreement are both high, with rationales emphasizing different experimental aspects indicating meaningful uncertainty. In Example 3 (Science), entropy is zero despite non-trivial human disagreement, showing that when model rationales are homogeneous, entropy may fail to detect latent disagreement. This reflects a known failure case: confident but brittle LLM justifications.

\paragraph{Summary.}
Overall, semantic entropy correlates with grader disagreement, rises with disagreement intensity, and generalizes across tasks. Its discriminative power is strongest in high-disagreement scenarios, and it captures underlying ambiguity in rubric interpretation. While not infallible, semantic entropy provides an interpretable signal for surfacing cases where model decisions may merit educator review. While the correlation between semantic entropy and grader disagreement is modest (r = 0.17), this effect is consistent and statistically robust. Even small predictive signals can be valuable as triage tools in educator workflows.

\section{RQ2: Is Semantic Entropy Effective Across Subjects?}

\subsection{Method}

Building on RQ1’s instance-level evaluation, RQ2 examines whether semantic entropy consistently reflects grader disagreement across different subject domains. Each response in the ASAP-SAS dataset is annotated with a subject label (e.g., Science, English, Biology), enabling stratified analysis.  We retain the same rationale generation protocol and entropy computation as in RQ1 to ensure consistency. Importantly, the prompt template remains domain-agnostic, no subject-specific tuning is applied allowing us to isolate the generalizability of semantic entropy from potential confounds like prompt optimization.

\subsection{Experiment}

To evaluate the subject-wise generalizability of semantic entropy, we partition the 2750 sampled responses into four subject areas based on the ASAP-SAS dataset: Science, English Language Arts (ELA), Biology, and English. Within each domain, we compute the Pearson and Spearman correlations between semantic entropy \(H(x_i)\) and human grader disagreement \(\delta_i\), and assess the ability of entropy to classify high-disagreement instances using binary accuracy. To further explore structural factors influencing entropy, we categorize tasks into two groups source-dependent and non-source-dependent depending on whether the student response is expected to incorporate information from an external source (e.g., a passage or experiment description or figures). This division enables us to examine whether the need for external grounding, which increases cognitive complexity and rubric ambiguity, also drives higher semantic entropy.

\subsection{Analysis and Findings}

\paragraph{Subject-Specific Predictive Performance.}  
Semantic entropy maintains predictive relevance across educational domains but exhibits substantial variation in effectiveness. Table~\ref{tab:subject_perf_correl} shows that entropy correlates moderately with human grader disagreement in interpretive subjects like {Biology} (\(r = 0.218, p = 2.28 \times 10^{-7}\)) and {English} (\(r = 0.206, p = 2.5 \times 10^{-9}\)), indicating that explanation variability aligns with rubric ambiguity in those domains. By contrast, correlations are weak and statistically insignificant in rubric-constrained or fact-based domains such as {Science} (\(r = -0.001, p = 0.988\)) and {English Language Arts (ELA)} (\(r = 0.010, p = 0.82\)). 

\begin{figure*}[!htb]
\centering
\scalebox{0.6}{
\includegraphics[width=\textwidth]{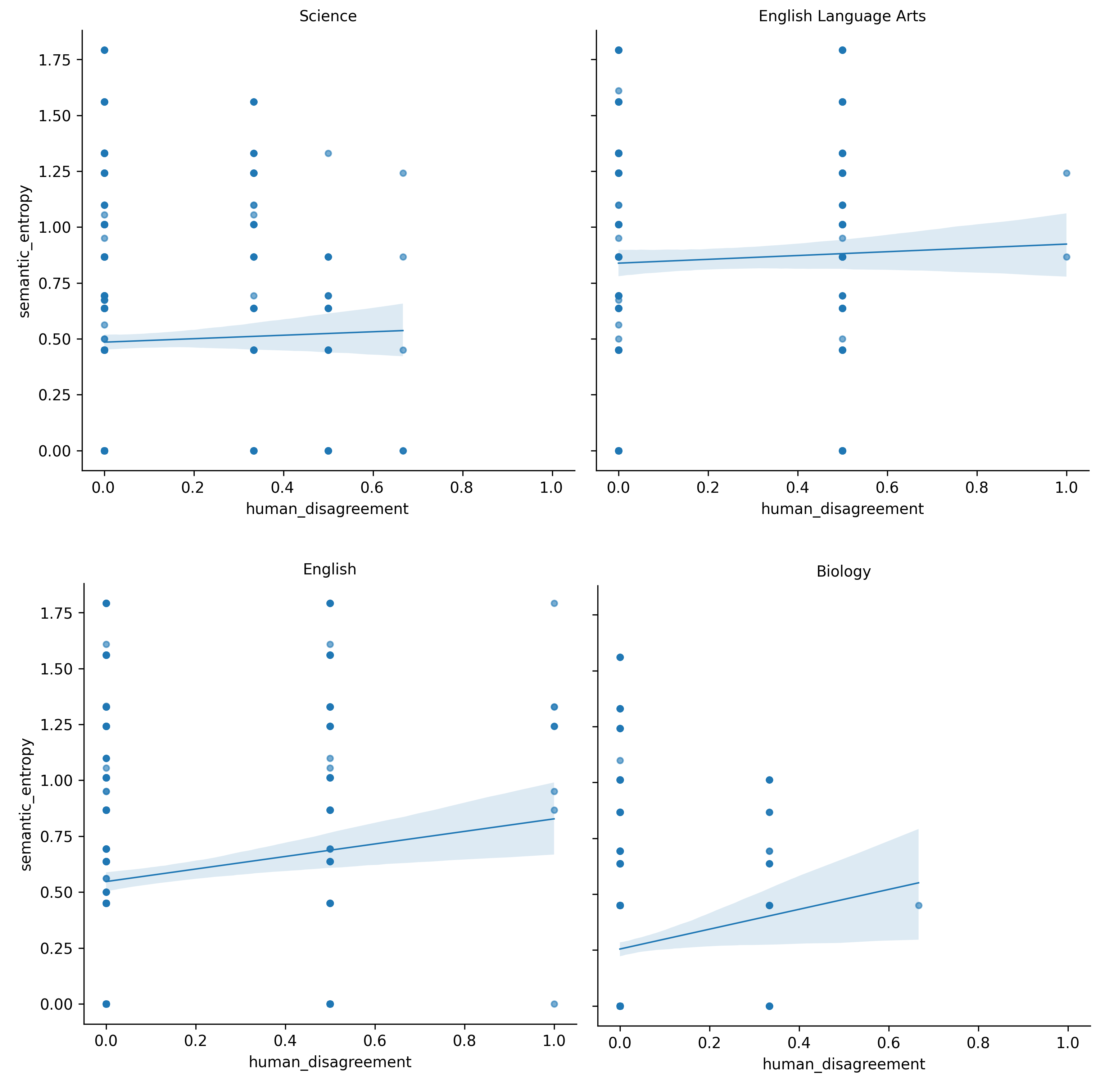}}
\caption{Scatterplots showing the relationship between semantic entropy and human score disagreement across subjects. Regression lines indicate fit with 95\% confidence intervals. Clearer trends are visible in Biology and English, while Science and ELA show weak or no association.}
\label{fig:subjectwise_entropy_scatter}
\end{figure*}

This trend is mirrored in classification accuracy results (Table~\ref{tab:subject_perf}), where semantic entropy yields its highest alignment with human graders in {Biology} (62.73\% and 65.45\%) and lowest in {Science} (48.24\% and 48.00\%). These findings are visually reinforced in Figure~\ref{fig:subjectwise_entropy_scatter}, which displays subject-specific scatterplots of semantic entropy against human disagreement. The positive slopes in Biology and English show consistent relationships between entropy and disagreement, while flat trends in Science and ELA support the statistical insignificance observed in correlation tests.

\begin{table}[!htb]
\centering
\caption{Subject-specific accuracy of average LLM score (Temperature = 1) in matching human scores. \textbf{Accuracy (Score1)} reflects exact match with the first human grader; \textbf{Accuracy (Score2)} reflects exact match with the second human grader.}
\label{tab:subject_perf}
\scalebox{0.75}{
\begin{tabular}{|c|c|c|c|}
\hline
\textbf{Subject} & \textbf{Count} & \textbf{Accuracy (Score1)} & \textbf{Accuracy (Score2)} \\
\hline
Science & 825 & 48.24\% & 48.00\% \\
English Language Arts & 550 & 51.09\% & 55.82\% \\
Biology & 550 & 62.73\% & 65.45\% \\
English & 825 & 48.61\% & 50.42\% \\
\hline
\textbf{Overall} & \textbf{2750} & \textbf{51.82\%} & \textbf{53.78\%} \\
\hline
\end{tabular}}
\end{table}

\begin{table*}[!htb]
\centering
\caption{Subject-specific correlation between semantic entropy and human score disagreement. Reported values include Pearson \(r\) and Spearman \(\rho\) with associated \(p\)-values.}
\label{tab:subject_perf_correl}
\scalebox{0.8}{
\begin{tabular}{|c|c|c|c|}
\hline
\textbf{Subject} & \textbf{Pearson \(r\) with \(p\)-value} & \textbf{Spearman \(\rho\) with \(p\)-value} & \textbf{Significance} \\
\hline
Biology & \( r=0.218\ (p=2.28\times10^{-7}) \) & \( \rho=0.225\ (p=9.64\times10^{-8}) \) & Moderate \\
English & \( r=0.206\ (p=2.5\times10^{-9}) \) & \( \rho=0.205\ (p=2.63\times10^{-9}) \) & Moderate \\
English Language Arts & \( r=0.010\ (p=0.82) \) & \( \rho=0.015\ (p=0.722) \) & Not significant \\
Science & \( r=-0.001\ (p=0.988) \) & \( \rho=-0.004\ (p=0.917) \) & Not significant \\
\hline
\end{tabular}}
\end{table*}

\paragraph{Statistical Validation of Subject Differences.}  
To test whether subject domain significantly influences entropy’s alignment with human disagreement, we conduct a Kruskal–Wallis test. The results reveal a highly significant difference across domains (\(H = 263.45, p = 8.04 \times 10^{-57}\)), with an effect size of \(\eta^2 = 0.085\). This indicates that domain characteristics explain roughly 8.5\% of the variance in entropy–disagreement correlation, substantiating the hypothesis that entropy is more informative in open-ended or interpretive tasks.

\paragraph{Task-Level Variation: Source Dependency.}  
We also note that some of the subject-level differences in entropy alignment may stem from task design. In particular, source-dependent tasks those requiring reference to external passages or visuals exhibit substantially higher mean entropy (\(H = 0.616\)) than non-source-dependent tasks (\(H = 0.264\)), with a large effect size (\(\Delta = +0.352\), \(p = 1.27 \times 10^{-44}\)). This suggests that interpretive load introduced by grounding may amplify rationale diversity and affect entropy–disagreement alignment across domains.

\paragraph{Summary.}  
In summary, while semantic entropy generalizes across educational subjects, its predictive strength varies by domain and task type. It proves most effective in tasks requiring interpretive reasoning, subjective rubric application, or external content integration settings that inherently admit more diverse justifications. These findings position semantic entropy as a domain-sensitive diagnostic signal, particularly well-suited for triaging complex or ambiguous grading cases.

\section{RQ3: Does Semantic Entropy Reflect Task Structure Specifically Source Dependency?}

\subsection{Method}

We examine whether semantic entropy is sensitive to task-level structural variation, particularly the presence of external grounding. In short-answer grading, \textit{source-dependent} tasks those that require reasoning over a reading passage, diagram, or other external material often introduce interpretive complexity and rubric ambiguity. These characteristics may increase the diversity of plausible explanations, even for the same student response.

The ASAP-SAS dataset includes a binary source-dependency flag for each prompt. We partition our 3,000-example sample into \textit{source-dependent} ($n = 2200$) and \textit{non-source-dependent} ($n = 550$) subsets. We reuse the entropy computation protocol from RQ1: for each student response, we generate six rationales using GPT-4, cluster them via bidirectional entailment, and compute semantic entropy over the resulting clusters. The model is not explicitly informed about source-dependence, ensuring that observed effects reflect differences in task structure rather than prompt variation.

\subsection{Experiment}

We compare entropy statistics and correlation with human grader disagreement between the two subsets. First, we compute mean entropy per group and evaluate statistical significance using the Mann–Whitney U test. Next, we assess alignment between entropy and disagreement $\delta_i$ within each subset using Pearson correlation. To control for confounds, we ensure comparable distributions across response length, rubric scales, and subject domains.

\subsection{Analysis and Findings}

\paragraph{Effect of Source Dependency on Entropy Magnitude.}
We find that source-dependent responses exhibit substantially higher semantic entropy (\( H = 0.551 \)) compared to non-source-dependent ones (\( H = 0.257 \)), with a large mean difference of \( \Delta = +0.294 \). This difference is statistically significant (Mann–Whitney U test, \( p = 1.27 \times 10^{-44} \)) and robust to variation in response length and domain composition. A mixed-effects regression confirms the main effect of source dependency on entropy (\( p = 0.030 \)), suggesting that external grounding reliably amplifies explanation diversity.

\paragraph{Entropy–Disagreement Alignment Across Task Types.}
We further observe that entropy correlates more strongly with human grader disagreement in source-dependent tasks (\( r = 0.218, p = 2.28 \times 10^{-7} \)) than in non-source-dependent ones (\( r = 0.109, p = 2.69 \times 10^{-7}\)). This indicates that semantic entropy not only increases in magnitude when grounding is present, but also becomes more predictive of human disagreement reinforcing its utility as a task-sensitive uncertainty signal.

\paragraph{Rubric Structure and Justification Space.}
To interpret these patterns, we examine representative rubric formulations across task types. High-entropy tasks often involve criteria such as “thoughtful response” or “deep analysis,” which allow multiple valid reasoning pathways. This rubric openness results in diverse AI-generated rationales, even when human graders agree on the score. In contrast, low-entropy tasks are associated with tightly constrained rubrics (e.g., “define,” “calculate”) that limit justification variability.

\paragraph{Summary.}
Our findings suggest that semantic entropy is not merely an artifact of model variability, but reflects the underlying interpretive space defined by task structure and rubric design. Based on these observations, we propose a preliminary rubric-based task taxonomy that serves as a hypothesis for future exploration: (i) \textbf{High Entropy Tasks} involve abstract interpretation, subjective evaluation, or synthesis (e.g., literary analysis, argumentative writing); (ii) \textbf{Medium Entropy Tasks} involve explanatory reasoning with partial grounding or ambiguity (e.g., inference-based STEM tasks); and (iii) \textbf{Low Entropy Tasks} emphasize factual recall or procedural correctness (e.g., definitions, computations). This taxonomy provides a potential lens for understanding how rubric openness and task complexity may influence explanation diversity. However, we caution that these interpretations are exploratory and subject to validation in follow-up work.

\section{Discussion}

Our study demonstrates that \textit{semantic entropy}, computed over clusters of model-generated rationales, serves as a reliable signal of epistemic uncertainty in automated short-answer grading. Across RQ1–RQ3, we find that entropy captures not just output variability, but underlying justification diversity often aligning with human grader disagreement, especially in contexts with interpretive openness or rubric ambiguity.

First, entropy scales with grader disagreement intensity and remains predictive across diverse prompts, highlighting its robustness as a disagreement proxy. Second, its effectiveness is modulated by subjects and rubric structure: tasks in interpretive subjects (e.g., Biology, English) exhibit stronger entropy–disagreement alignment, while factual or rigidly scored domains (e.g., Science, ELA) show weaker effects. Finally, task structure plays a crucial role: source-dependent questions consistently produce higher entropy, reflecting the cognitive complexity of integrating external context and supporting multiple valid reasoning paths.

While the proposed task taxonomy highlights plausible relationships between rubric structure and explanation diversity, we acknowledge several potential confounds. Entropy variation may also arise from factors such as student response length, rubric wording, prompt phrasing, or even sampling-induced hallucinations by the language model. Therefore, we treat the taxonomy as an exploratory hypothesis rather than a definitive classification of task complexity. Future work should conduct controlled studies to isolate these variables and validate whether entropy-based categorization reflects intrinsic cognitive demand or structural ambiguity.

Together, these findings position semantic entropy as a task-sensitive signal that supports triage in educational AI systems. By highlighting responses likely to admit multiple valid interpretations even when scores agree entropy offers a principled mechanism for surfacing cases where human review is warranted or rubric refinement may be necessary. To support real-world educator workflows, we propose a practical quadrant-based decision strategy that combines semantic entropy and human grader disagreement. This framework helps determine when AI-generated scores should be trusted versus flagged for review:

\begin{itemize}
    \item \textbf{High entropy + high disagreement:} Flag for mandatory human review or rubric revision. These are cases where both the model and humans show uncertainty, indicating a need for careful examination.
    \item \textbf{High entropy + low disagreement:} Indicates rubric underspecification or hidden response diversity. Even when humans agree, diverse AI justifications may reveal ambiguity in the rubric or multiple valid reasoning paths.
    \item \textbf{Low entropy + high disagreement:} Suggests model overconfidence or grader inconsistency. The model produces consistent explanations, yet humans diverge - this may flag a risk in automated scoring.
    \item \textbf{Low entropy + low disagreement:} Safe candidates for full automation. Both human and model signals are consistent, suggesting reliable grading decisions.
\end{itemize}

\section{Conclusion and Future Work}
In this work, we proposed semantic entropy as a principled, explanation-level signal for capturing epistemic uncertainty in AI-assisted short-answer grading. By clustering GPT-4-generated rationales and measuring their semantic divergence, our method offers an interpretable alternative to conventional scalar-based uncertainty metrics. Our experiments on the ASAP-SAS dataset demonstrate that semantic entropy aligns with human grader disagreement (RQ1), generalizes across academic domains with varying rubric structures (RQ2), and reflects cognitive task complexity, particularly in source-dependent assessments (RQ3). These results underscore semantic entropy’s potential as a domain- and task-sensitive signal for triaging ambiguous or contentious grading cases in educational settings.

While our findings are encouraging, several limitations remain that point to promising avenues for future work. First, our current pipeline uses GPT-4 for both generating rationales and performing entailment-based clustering. This raises concerns about bias and methodological circularity, which we plan to address by decoupling the generation and clustering components using independent models. Second, we acknowledge that some degree of entropy may be attributed to stochastic sampling noise rather than genuine epistemic uncertainty. Future experiments will explore robustness across temperature settings and sampling strategies to better distinguish between noise and substantive variability in explanations.

Additionally, our approach does not yet benchmark against simpler uncertainty measures such as score variance, response length, or softmax confidence. Comparing semantic entropy with these baselines will help clarify its unique contributions. Another key direction is the validation of our rationale clusters through human annotation studies, to determine whether the observed variability reflects distinct rubric aligned reasoning paths or superficial linguistic variation. This will also help assess the interpretability and pedagogical relevance of entropy as an educational signal.

Finally, we envision deploying semantic entropy in real-time educational systems where it can support educators by highlighting student responses that warrant manual review or rubric refinement. Integrating entropy into intelligent tutoring workflows such as hint generation, adaptive scaffolding, or teacher dashboards could help build more transparent and trustworthy AI-assisted learning environments.

% \bibliography{sample-ceur}

%% The declaration on generative AI comes in effect
%% https://ceur-ws.org/GenAI/Policy.html
\section*{Declaration on Generative AI}
During the preparation of this work, the author(s) used ChatGPT, Grammarly in order to: Grammar and spelling check, Paraphrase and reword. After using this tool/service, the author(s) reviewed and edited the content as needed and take(s) full responsibility for the publication’s content.

%%
%% Define the bibliography file to be used
\bibliography{sample-ceur}

\appendix


% The sources for the ceur-art style are available via
% \begin{itemize}
% \item \href{https://github.com/yamadharma/ceurart}{GitHub},
% % \item \href{https://www.overleaf.com/project/5e76702c4acae70001d3bc87}{Overleaf},
% \item
%   \href{https://www.overleaf.com/latex/templates/template-for-submissions-to-ceur-workshop-proceedings-ceur-ws-dot-org/pkfscdkgkhcq}{Overleaf
%     template}.
% \end{itemize}

\end{document}